\documentclass[12pt,onecolumn]{article}
 \usepackage{mathtools}

\usepackage{amsmath}  
 \usepackage{amssymb}
\usepackage{amsthm}
\usepackage[utf8x]{inputenc}
\usepackage{authblk}
\usepackage{url}
\usepackage{graphicx}
\usepackage{tikz,tkz-fct}

\usepackage{caption}
\usepackage{subcaption}
\usepackage{setspace}
\usetikzlibrary{plotmarks}
\usepackage{pgfplots}

\usepackage{fullpage}
\usepackage{times}
\usepackage{helvet}
\usepackage{courier}
\usepackage{paralist}
\usepackage{latexsym}
\usepackage{url}
\usepackage[all]{xy}
\usepackage{amsmath}
\usepackage{amssymb}
\usepackage{comment}
\usepackage{moresize}
\usepackage{xcolor}
\usepackage{pifont}
\usepackage{times}
\usepackage{graphicx}
\usepackage{savesym}
\savesymbol{AND}
\usepackage{xspace}
\usepackage[textsize=tiny]{todonotes}
\usepackage{tikz}
\usepackage{pgfplots}
\usepackage{pgf}
\usepackage{algorithm}
\usepackage{algorithmic}
\usepackage{xspace}
\usepackage{comment}
\usepackage{footnote}
\usepackage{placeins}
\usepackage{cleveref}
\usetikzlibrary{intersections}
\usetikzlibrary{arrows,calc,fit,patterns,plotmarks,shapes.geometric,shapes.misc,shapes.symbols,   shapes.arrows,   shapes.callouts,   shapes.multipart,   shapes.gates.logic.US,   shapes.gates.logic.IEC,   er,   automata,   backgrounds,   chains,   topaths,   trees,   petri,   mindmap,   matrix,   calendar,   folding, fadings,   through,   positioning,   scopes,   decorations.fractals,   decorations.shapes,   decorations.text,   decorations.pathmorphing,   decorations.pathreplacing,   decorations.footprints,   decorations.markings, shadows,circuits}
\tikzstyle{decision}=[diamond,draw]
\tikzstyle{line}=[draw]
\tikzstyle{elli}=[draw,ellipse]
\tikzstyle{arrow} = [thick]
\newcommand{\eqdef}{\stackrel{\Delta}{=}}
\newcommand{\hQ}{\hat{Q}}
\newcommand{\D}{\mathcal{D}}

\newcommand{\hQp}{\hat{Q}^{\pi}}
\newcommand{\mb}{\mbox{ }}

\newcommand{\R}{\mathrm{R}}

\newcommand{\ra}{\rightarrow}

\newcommand{\Ls}{\mathcal{L}}

\newcommand{\Ps}{\mathcal{P}}

\newcommand{\M}{\mathcal{M}}
\newcommand{\K}{\mathcal{K}}

\newtheorem{theorem}{Theorem}

\newtheorem{example}{Example}

\newcounter{subequation}[equation]

\def\mathdisplay#1{%
  \ifmmode \@badmath
  \else
    $$\def\@currenvir{#1}%
    \let\dspbrk@context\z@
    \let\tag\tag@in@display \SK@equationtrue 
    \global\let\df@label\@empty \global\let\df@tag\@empty
    \global\tag@false
    \let\mathdisplay@push\mathdisplay@@push
    \let\mathdisplay@pop\mathdisplay@@pop
    \if@fleqn
      \edef\restore@hfuzz{\hfuzz\the\hfuzz\relax}%
      \hfuzz\maxdimen
      \setbox\z@\hbox to\displaywidth\bgroup
        \let\split@warning\relax \restore@hfuzz
        \everymath\@emptytoks \m@th $\displaystyle
    \fi
}

\newcounter{algostep}

\newcounter{acalgorithm}

\title{Shaping Proto-Value Functions via Rewards}
\author{Chandrashekar Lakshmi Narayanan}
\author{Raj Kumar Maity}
\author{Shalabh Bhatnagar}

\affil{Dept of Computer Science and Automation, Indian Institute of Science}

\date{}
\allowdisplaybreaks

\pgfplotsset{
tick label style={font=\large},
}

\begin{document}
\maketitle
\begin{abstract}
Learning value function is an important sub-problem in solving a given reinforcement learning task. The choice of representation for the value function directly affects learning. The most widely used representation for the value function is the linear architecture, wherein, the value function is written as a linear combination of a `pre-selected' set of basis functions. In such a scenario, choosing the right basis function is crucial in achieving success. Often, the basis functions are either selected in an ad-hoc manner or their choice is based on the domain knowledge that is specific to the given RL task. However, it is desirable to be able to choose the basis functions in a task-independent manner. The \emph{proto-value} functions (PVFs) are task-independent basis functions and are based on the topology of the state space. Being eigen functions of the random walk operator, the proto-value functions capture the connectivity and neighborhood information. \par
In contrast to supervised learning, agent performing an RL task needs to learn from the rewards. However, in goal-based RL tasks, the rewards are delayed, i.e., the agents receive feedback only after reaching the goal state and such delay can cause poor learning rates. Reward shaping is the mechanism of providing additional rewards for correct behavior in non-goal states, thereby aiding the learning process.\par
In this paper, we combine task-dependent reward shaping and task-independent proto-value functions to obtain reward dependent proto-value functions (RPVFs). In constructing the RPVFs we are making use of the immediate rewards which are avaialble during the sampling phase but are not used in the PVF construction. We show via experiments that learning with an RPVF based representation is better than learning with just reward shaping or PVFs. In particular, when the state space is symmetrical and the rewards are asymmetrical, the RPVF capture the asymmetry better than the PVFs.
\end{abstract}

\section{Introduction}
Reinforcement Learning (RL) tasks problems are cast in the framework of Markov decision processes (MDPs). In the MDP setting, dynamics of the underlying environment evolves within a set of states called the state-space, and the agent performs actions to control the state of the system. The agent receives a reward which is dependent on the state and the action it performs. The agent aims to maximize the discounted infinite sum of the rewards obtained as a result of its actions. Formally, any action selection mechanism is known as a policy and the agent aims to learn the optimal policy. \par
In order to learn the optimal behavior/policy, the agent first needs to evaluate the current behavior. The value function $J_u$ corresponding to a given policy $u$, is a map from the state space to real numbers, and captures the total discounted reward that agent collects by following the policy $u$. From the knowledge of the value function, the agent can improve its behavior and hence learning the value function in an efficient manner assumes importance.\par
Agent's choice for representing the value function affects the learning process.  A desirable property is that the representation has to be \emph{compact} (i.e., it should be easy to compute and store the value function). The linear function representation is the most widely used, wherein, the value function is represented as a linear combination of the basis functions. In general, the choice of the basis is guided by the task-specific knowledge and when there is no such task-specific information, primitive functions such as radial basis functions, polynomial bases and tile coded bases are chosen. Neverthless, it is desirable to be able to construct basis functions with little or no information about the task.\par
The proto-value functions (PVFs) \cite{proto} are bases that can be constructed in task-independent manner, and have been applied to a wide variety of domains. The PVFs are obtained by diagonalizing symmetric diffusion operators on an empirically learned graph representing the underlying state space. A diffusion model or the random walk on an undirected graph, where the probability of transitioning from a vertex (state) to its neighbor is proportional to its degree, is intended to capture information flow on a graph. The PVFs being equivalent to Fourier bases were shown to capture the intricate connectivity in the underlying state space which primitive bases such as Fourier or polynomials do not capture. \cite{icmlproto} presented the representational policy iteration (RPI) algorithm by combining the PVF basis construction with least squares policy iteration (LSPI).
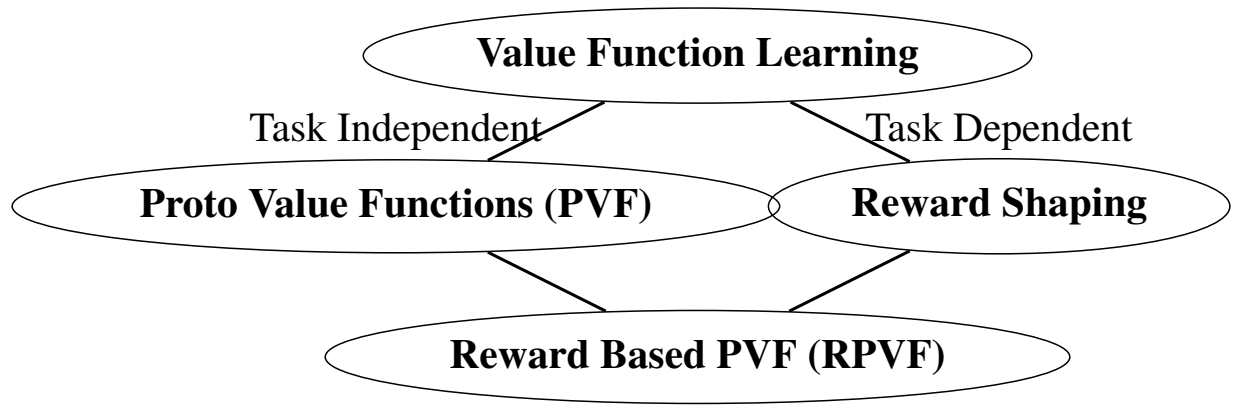
\begin{figure}[h!]
\resizebox{\columnwidth}{!}{%
\tikzstyle{every node} = [align=center]
\begin{tikzpicture}[domain=0:7.7,scale=0.7,font=\small,axis/.style={very thick, ->, >=stealth'}]
\node at ($(0,0)$) (MDP) [elli]{\textbf{Value Function Learning}};
\node at ($(MDP)+(-4,-2)$)(DP) [elli]{\textbf{Proto Value Functions (PVF)}};
\node at ($(MDP)+(4,-2)$)(RS) [elli]{\textbf{Reward Shaping}};
\node at ($(MDP)+(0,-4)$)(RL) [elli]{\textbf{Reward Based PVF (RPVF)}};
\node at ($(MDP)+(-4,-1)$)() {Task Independent};
\node at ($(MDP)+(4,-1)$)() {Task Dependent};
\draw [arrow] (MDP) -- (DP) -- (RL);
\draw [arrow] (MDP) -- (RS) -- (RL);
\end{tikzpicture}
}
\caption{Shows the idea behind reward based proto-value functions}
\end{figure}
\par
Whilst topology dictated by the underlying graph of the MDP constitutes one form of domain knowledge (feature selection), researchers have also looked at other means to enable faster learning. Reward shaping \cite{rs1,rs2,rs3,rs4,rsng,rspt} is the process of providing additional rewards to the learning agent to guide its learning process. The reward function has to be chosen in such a way that it preserves the optimal policy. Reward shaping is task-specific since the shaping function is not dependent on the state space and varies depending on the goal or the reward structure.\\
\textbf{Our Contribution} In this paper, we combine the ideas of task-independent PVF construction and the task-specific reward shaping to construct Reward based Proto-Value Functions (RPVFs). The idea behind such a construction is the observation that the actual neighborhood we are interested in is the one that is generated by the value function itself. While, topologically near states might have similar values, it is also true that the value is affected by the immediate rewards. In a general reward MDP, though the immediate rewards are obtained during the sampling phase, they are not used in the PVF construction. 
We modify the diffusion operator using the immediate rewards in order to construct the RPVFs. We show success of RPVFs in experiments on benchmark RL tasks.\\
\textbf{Highlights} Our experiments demonstrate that similarity matrices other than the diffusion matrix can be used to generate features, and that reward shaping does benifit when the features are ill chosen. In particular, when the state space is symmetrical and the rewards are asymmetrical, the RPVF capture the asymmetry better than the PVFs.\\
\textbf{Organization} We first present the overview of the reinforcement learning paradigm emphasizing the need to learn the value function. We then discuss proto-value functions in brief. Next we present the RPI algorithm following which we discuss the objective of reward shaping. Finally, we present RPVF and experimental results.

\section{Reinforcement Learning Paradigm}
\begin{figure}[h!]
\centering
\tikzstyle{every node} = [align=center]
\begin{tikzpicture}[domain=0:7.7,scale=0.7,font=\small,axis/.style={very thick, ->, >=stealth'}]

\draw [thick](-4.500,-2.50)rectangle(5.00,-1.50);
\node at (-0.00,-2.00){\textbf{Environment}};

\draw [thick](-4.500,-0.10)rectangle(5.00,3.00);
\node at (-0.00,2.75) {\textbf{RL Agent}};

\node at ($(-2.5,1.5)$) (MDP) [elli]{Samples};
\node at ($(2,2)$) (MDP) [elli]{Representation};
\node at ($(2,0.75)$) (MDP) [elli]{Learning Algorithm};

\draw [thick,->] (-1,-1.5)--(-2.5,1);
\node at (-3, -0.5) {$(s_t,r_t)$};
\node at (4.5, -0.5) {$(a_t)$};
\draw [thick,->] (4.2,0.4)--(3.5,-1.5);


\end{tikzpicture}
\label{setting}
\caption{Various blocks in the RL paradigm}
\end{figure}
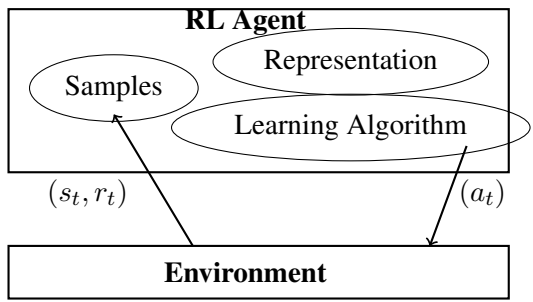
\textbf{Environment} The dynamics of the underlying environment can be captured in the framework of Markov decision process (MDP). An MDP is a $4$-tuple $<S,A,P,R>$, where $S$ is the state space, $A$ is the action space, $P$ is the probability transition kernel and $R$ is the reward function. The probability transition kernel $P$ specifies the probability $p_a(s,s')$ of transitioning from state $s$ to state $s'$ under the action $a$. The reward function $R$ is a map $R\colon S\times A \ra \R$ that specifies the reward obtained for performing action $a\in A$ in state $s\in S$ and is denoted by $r_a(s)$.\\
\textbf{Agent Behavior} The behavior of the agent is captured by the way the actions it makes in each and every state. In the MDP parlance, this action selection mechanism is called the policy. Formally, by a policy, we mean a sequence $\mu=\{\mu_0,\ldots,\mu_n,\ldots\}$ of functions $\mu_i, i\geq 0$ that describe the manner in which an action is picked in a given state at time $i$. 
Two important types of policies that are also useful are: $(i)$ \emph{Stationary Randomized Policy} (SRP), given by $\mu=\{\mu_0,\ldots,\mu_i,\ldots\}$, where $\mu_i\equiv \pi,\mb\forall i\geq 0$ with $\pi(s,\cdot)$ being a probability distribution over the set of  actions for any $s\in S$. $(ii)$ \emph{Stationary Deterministic Policy} (SDP), given by $\mu=\{\mu_0,\ldots,\mu_i,\ldots\}$, where $\mu_i\equiv u,\mb\forall i\geq 0$ with $u\colon S\ra A$ being a map from the state space to the action space.\par
Note that an SDP is trivially a SRP as well. By abuse of notation, we refer an SRP by $\pi$ and an SDP by $u$.
Further, under a stationary policy $u$ (or $\pi$), the MDP is a Markov chain and we denote its probability transition kernel by $P_u=(p_{u(i)}(i,j),i,j=1,\ldots,n)$ (or $P_\pi=(p_{\pi(i)}(i,j),i,j=1,\ldots,n)$, where $p_{\pi(i)}(i,j)=\sum_{a\in A}\pi(i,a)p_a(i,j)$ and $\pi(i)=(\pi(i,a), a\in A)$).\par
\textbf{Value Function} We define the infinite horizon discounted reward value function under an SRP $\pi$ as $J^{\pi}(s)= E\bigg[ \sum^\infty_{t=0 }\alpha^t r_t|s_0=s,\pi \bigg ],$ where $\alpha \in (0,1)$ is the discount factor and $r_t=r_{a_t}(s_t)$ with $a_t\sim \pi(s_t,\cdot), \forall t\geq 1$. Similarly, we also define the infinite horizon discounted reward state-action value function under an SRP $\pi$ as $Q^{\pi}(s,a)= E\bigg[ \sum^\infty_{t=0 }\alpha^t r_t|s_0=s,a_0=a,\pi \bigg ]$.
\par
The optimal policy\footnote{In the infinite horizon discounted reward setting that we consider, one can find an SDP that is optimal \cite{BertB,Puter}} and the optimal value function obey the Bellman equation (BE) given below: $\forall s \in S$,
\begin{align}\label{bell}
\begin{split}
Q^*(s,a)&=\big(r_a(s)+\alpha \sum_{s'}p_a(s,s')\max_{a'\in A}Q^*(s',a')\big),\\
u^*(s)&=\arg\max_{ a\in A}Q^*(s,a).
\end{split}
\end{align}

\textbf{RL Agent} 
Any RL agent has three important building blocks or sub-functions namely sample collection, the representation and the learning algorithm. Learner represents state of the environment $s_t$ at time $t$ as a point in the feature space. Learning algorithm makes use of samples $(s_t,r_t),t\leq n$ obtained from the environment and its own past behavior $a_t,t\leq n$to learn.\\
The behavior of the agent is dictated by policy $\pi$ it makes use to choose the actions. From \eqref{bell} it is clear that in order to compute the optimal behavior $u^*$, the agent needs to learn $Q^*$. Even in the case when agent wants to improve a given policy $\pi$, it has to evaluate $Q^\pi$ and then substituting $Q^\pi$ in \eqref{bell} will lead to an improved policy \cite{BertB}. Thus, learning the value function is central in learning the correct behavior. Such learning is dependent on the agent's way of representing the value functions.\\
\textbf{Value Function Representation} The most widely used representation is the linear function representation, wherein, the value $Q^\pi(s,a)$ of state-action pair $(s,a)$ is expressed as a weighted combination of the feature corresponding to that state, i.e., $Q^\pi(s,a)=\sum_{i=1}^k \phi_i(s,a)^\top w^\pi (i)$, where $(\phi_i(s,a), i=1,\ldots,k)\in \R^k$ is the feature of the state $s$ and $w^\pi \in \R^k$ is a learned weight vector. Any linear representation can be compactly represented by its feature matrix $\Phi=[\phi_1|\ldots,\phi_k]$, where $\phi_i \in \R^{\mathcal{C}},i=1\ldots,k (\mathcal{C}=|S||A|)$ are the $k$ basis functions.\par
Classical numerical schemes such as value iteration, policy iteration and linear programming choose a look up table representation. Under the look up table representation the standard basis is chosen, i.e., $\phi_1=(1,0,\ldots,0)^\top$. Thus there are as many basis functions as number of state-action pairs and as a result they might not always be efficient.\\
\textbf{Learning Algorithm} 
The least squares policy iteration (LSPI) algorithm is a widely used algorithm that makes use of a linear representation to learn the value function.
LSPI \cite{lspi} makes use of least squares temporal difference learning (LSTD) \cite{lstd} which computes $\hQp=\overset{k}{\underset{i=1}{\sum}} \phi_i w^\pi_i$ by solving an approximate fixed-point equation given by $
\Phi w^\pi\approx R+\alpha H_\pi \hQp$, 
where $H_\pi$ is the $\mathcal{C}\times \mathcal{C}$ matrix specifying the probability of transitioning between state-action pairs. On re-arranging we have
$\Phi w^\pi-\alpha H_\pi \Phi w^\pi\approx R$.\par
From least-squares regression we know that $w^\pi=(\Phi^\top D^\pi(\Phi w^\pi-\alpha H_\pi))^{-1}\Phi^\top D^\pi R$, where $D^\pi$ is a diagonal matrix whose entries are the stationary distributions of the various state-action pairs under the SRP $\pi$. Further, by letting $A^\pi=(\Phi w^\pi-\alpha H_\pi)$ and $b^\pi=\Phi^\top D^\pi R$, it follows that $w^\pi=({A^\pi})^{-1}b^\pi$.\par
In the section to follow, we will describe the proto-value functions, and the representational policy iteration algorithm (RPI). The RPI uses PVFs in the LSPI algorithm to learn the value function.

\section{Proto-Value Functions}
It is in general a good idea to select basis functions by using domain knowledge. When the domain knowledge is absent, representations based on well known primitive functions such as radial, polynomial or Fourier can be used. However, such representations based on the primitive functions might not yield good results.Hence, it is desirable to be able to choose the basis functions in a task-independent manner. 
The \emph{proto-value} functions are task-independent basis functions and are based on the topology of the state space. Being eigen functions of the random walk operator, the proto-value functions capture the connectivity/neighborhood information. We observe that such neighborhood information is also affected by the reward structure.\par
Let $G=(E,V)$ denote a graph with edge set $E$ and the vertex set $V$. Let $A=(A_{ij},i,j=1,\ldots,|V|,)$ denote the adjacency matrix, with $A_{ij}=1$ when vertices $i$ and $j$ are connected, and $A_{ij}=0$ when $i$ and $j$ are not connected.
 We now define the following matrices
\begin{center}
\FloatBarrier
\begin{table}[H]
\begin{tabular}{|c|l|}\hline
$A$& Adjacency Matrix\\\hline
$D$& Diagonal matrix with entries\\ 
& as row sums of $A$\\ \hline
$L=D-A$& Combinatorial Laplacian\\ \hline
$\Ls=D^{-1/2}L D^{-1/2}$& Normalized Laplacian\\ \hline
$W=D^{-1}A$& Random walk diffusion matrix\\ \hline
\end{tabular}
\end{table}
\end{center}
\vspace{-20pt}
In a graphical representation where vertices are the states of a system, the adjacency matrix $A$ can be treated as the measure of similarity and the eigen vectors  can be chosen for the representation of the basis function. For instance, the spectral clustering technique uses eigen-value (spectrum) of the similarity matrix to perform dimensionality reduction. One of the well used methods is choosing the eigen vector corresponding to the second smallest eigen- value of the (symmetric) normalized Laplacian.
A spectrally similar and alternative way is to choose eigen vector corresponding to the highest eigen-values of the random walk diffusion matrix $W=D^{-1}A $ which represent the transition probability from a vertex to its  neighbor vertex that is proportional to its degree. To see this note that $I-\mathcal{L}  = D^{-\frac{1}{2}}A D^{-\frac{1}{2}}= D^{-1}A$.\par
A look at the following expression of the value function $J_\pi$ throws light into why the diffusion matrix is helpful.
\begin{align*}
J^\pi=(I+\alpha P_\pi+\alpha^2 P_\pi^2+\ldots) R.
\end{align*}
Assuming that the transition matrix $P_\pi$ is diagonalizable, i.e., $P_\pi=\Phi^\pi (\Lambda) {\Phi^\pi}^\top=\sum_{i=1}^n\lambda_i (\phi^\pi_i){\phi_i^\pi}^\top$, the above expansion then becomes
\begin{align*}
J^\pi&=(I+\alpha\Phi^\pi (\Lambda) {\Phi^\pi}^\top+ \alpha^2\Phi^\pi (\Lambda^2) {\Phi^\pi}^\top)R\\
&=\sum_{i=1}^n \frac{1}{1-\alpha\lambda^\pi_i}\phi^\pi_i{\phi^\pi_i}^\top R\approx\sum_{i=1}^k \frac{1}{1-\alpha\lambda^\pi_i}\phi^\pi_i{\phi^\pi_i}^\top R
\end{align*}
where $i=1,\ldots,k$ are such that $\lambda^\pi_i, i=1,\ldots,k$ are the largest eigen values. Thus the value function can be approximated as a linear combination of the eigen vectors corresponding to the largest eigen values of the transition matrix (since $\alpha$ is fixed, $\frac{1}{1-\alpha\lambda}$ is higher for higher values of $\lambda$). However, in the absence of knowledge of the transition matrix $P_\pi$, one can make use of the diffusion matrix $W$ obtained from the graph adjacency matrix.
\begin{figure}[h!]
\centering
\begin{tikzpicture}[domain=0:7.7,scale=0.7,font=\small,axis/.style={very thick, ->, >=stealth'}]
\draw [thick](-4.00,-2.50)rectangle(-2.00,0.00);
\node at (-3.80,-0.25){\textbf{S}};
\draw [thick](-2.00,-2.50)rectangle(0.00,0.00);
\draw [thick](-0.00,-2.50)rectangle(2.00,0.00);
\node at (1.80,-2.30){\textbf{G}};
\draw (-2.2,-1.00)--(-1.8,-1.00);
\draw (-2.2,-1.20)--(-1.8,-1.20);

\draw (-0.2,-1.00)--(0.2,-1.00);
\draw (-0.2,-1.20)--(0.2,-1.20);

\node at (-4.20,-1.70){\textbf{21}};
\node at (-3.80,0.20){\textbf{20}};
\end{tikzpicture}
\caption{Three Room Task. Here the agent needs to start from $\textbf{S}$ and reach the goal-state $\textbf{G}$. Each room is of size $20\times 21$ and the walls that separate the room cause discontinuities and makes the representation based on primitive functions ineffective.}
\label{approx}
\end{figure}
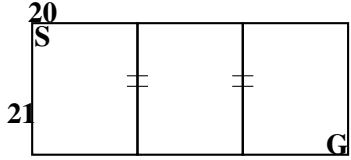
Consider the three-room problem \cite{icmlproto}, wherein, the agent has to move from the starting position $\mathbf{S}$ in the top-left side of the first room to the goal state $\mathbf{G}$ in the bottom-right of the third room. The task is particularly difficult because of the presence of the walls, which create a discontinuity, i.e., any representation based on primitive functions such as polynomial or radial would not account for the discontinuity. In \cite{icmlproto}, the authors demonstrated power of the proto-value functions in approximating such a complicated value function.

\vspace{-10pt}
\section{Representational Policy Iteration (RPI)}
The RPI algorithm \cite{icmlproto} will be the template algorithm that we will be using for our experiments.
Since the model information is not available, the LSTD algorithm learns it from the samples trajectories. We now present the LSPI algorithm (see Algorithm~\ref{alglspi}) which makes use of LSTDQ (see Algorithm~\ref{alglstdq}), a variant of the LSTD algorithm).
\FloatBarrier 
\begin{algorithm}[H] 
\caption{Representational Policy Iteration $(\D,\alpha,\Theta,k,\pi_0)$ }
\begin{algorithmic}[1]
\STATE Choose feature matrix $\Phi$ to be the top $k$ eigen-vectors of $\Theta$.
\FOR{$i= 0, 1, 2,\ldots,t-1$} 
\STATE Policy Evaluation Step: $w^{\pi_i}=LSTDQ(D,\pi_i)$
\STATE Policy Improvement Step: Set $\pi_{i+1}(s)=\underset{a \in A}{\arg\max}(\hQ^{\pi_i}(s,a)), \forall s\in S$.
\ENDFOR
\STATE Return $\pi_{\Theta}\eqdef\pi_{t}$.
\end{algorithmic}
\label{alglspi}
\end{algorithm}
\vspace{-10pt}
\FloatBarrier 
\begin{algorithm}[H]
\caption{$LSTDQ(D,\pi_i)$}
\begin{algorithmic}[1]
\STATE Initialize a policy $A_0=\mathbf{0}$, $b_0=\mathbf{0}$.
\FOR{$i= 0, 1, 2,\ldots,T$} 
\STATE $A^\pi_{i+1}=A^\pi_{i}+\phi(s_i,a_i)(\phi(s_i,a_i)-\alpha\phi(s'_i,\pi(s'_i))^\top$.
\STATE $b^\pi_{i+1}=b^\pi_{i}+\phi(s_i,a_i)r_{a_i}(s_{i+1})$.
\ENDFOR
\RETURN $w^\pi=({A^\pi})^{-1}b^\pi$.
\end{algorithmic}
\label{alglstdq}
\end{algorithm}
Here $\D$ is the sampled data, $\Theta$ is the matrix whose $k$ eigen vectors are used as features, $\pi_0$ is the initial policy, $t$ and $T$ are integers which are chosen large enough to ensure convergence.

\section{Reward Shaping}
The most fundamental difference between reinforcement learning (RL) tasks and supervised learning is that, in RL, the \emph{agent} needs to learn using the \emph{feedback} obtained in the form of the rewards it receives for its actions. Such learning via feedback makes RL tasks more challenging than \emph{supervised} learning problems wherein the \emph{correct/right} actions are provided to the learner in the training stage of the problem. This difficulty of learning from feedback is pronounced especially in the case of goal based tasks, wherein, the agent has to reach the \emph{goal-state} from any part of the state space, however, the agent receives no reward at all in states other than the goal-state. Thus the behavior in the states other than the goal-state is not clear, which results in slower convergence of the RL algorithms. In such a scenario, rewarding the correct behavior of the agents in the intermediate states can be helpful. This mechanism of providing external rewards for right behavior in addition to the rewards obtained from the environment is called \emph{reward shaping}. Reward shaping serves as indicator of the agent's progress, and can be seen as an improvement to the algorithmic part of the learning agent. 
\par
Reward-shaping was first introduced in \cite{rsng}, wherein, the authors furnished the conditions under which reward shaping preserves the optimal policies.  In particular, it is known that reward shaping functions $R'$ that are potential functions as well preserve the structure (see \Cref{th}).
\begin{theorem}[Theorem~$1$ of \cite{rsng}]\label{th}
Given an MDP $\M=\{S,A,P,R\}$ and a reward shaping function $R'\colon S\times A \times S \ra \R$, the MDP $\M'=\{S,A,P,R+R'\}$ has the same optimal policies as $\M$ iff $R'$ is \emph{potential} based, i.e.,
\begin{align}
R'(s,a,s')=\alpha \psi(s')-\psi(s),
\end{align}
\end{theorem}
for some $\psi\colon S \ra \R$.

\section{Proto Value function shaping using rewards}
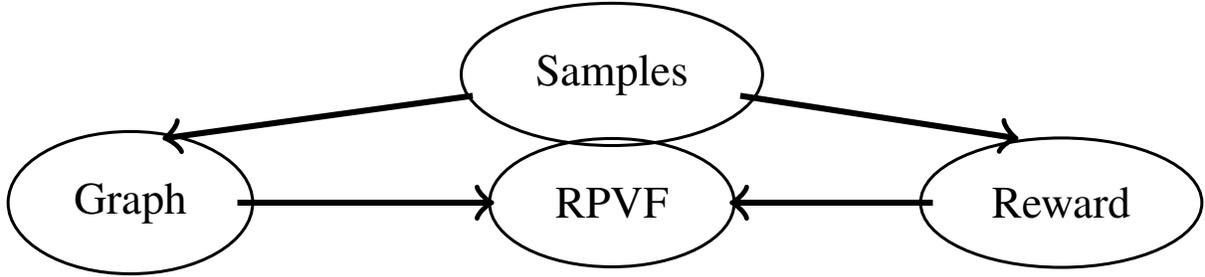
\begin{figure}[h!]
\centering
\tikzstyle{every node} = [align=center]
\resizebox{\columnwidth}{!}{
\begin{tikzpicture}[domain=0:7.7,scale=0.5,font=\tiny,axis/.style={very thick, ->, >=stealth'}]
\node at ($(1,2.7)$) (MDP) [elli]{Samples};
\node at ($(-3.5,1.5)$) (MDP) [elli]{Graph};
\node at ($(1,1.5)$) (MDP) [elli]{RPVF};
\node at ($(5.2,1.5)$) (MDP) [elli]{Reward};
\draw [thick,->] (-2.5,1.5)--(-.1,1.5);
\draw [thick,->] (4,1.5)--(2.1,1.5);
\draw [thick,->] (-0.3,2.5)--(-3.2,2.1);
\draw [thick,->] (2.2,2.5)--(4.8,2.1);
\end{tikzpicture}
}
\label{setting}
\caption{Construction of RPVF using connectivity as well as rewards}
\end{figure}
The PVFs as well as reward shaping, though conceptually different, ultimately help in efficient value function learning. PVFs capture the underlying neighborhood information by making use of the connectivity in the graph associated with the MDP. However, in reality what we care about is not the nearness associated with the topological neighborhood in state space but the nearness of the value functions. Such nearness, we observe, is also affected by the underlying reward structure. Also, PVFs are constructed by sampling the state space, a phase during which we also get to observe the immediate rewards. While in the case of goal oriented tasks, the immediate rewards are $0$, it might not hold true for MDPs with a general reward structure.\par 
Immediate rewards are indicators of agent's preference and actions locally. Consider for instance a goal-based MDP, however, with negative rewards for certain states. Given that the agent needs to move a step closer to the goal at each stage and that the states with negative rewards are equivalent to making additional steps, the agent's immediate action will be to prefer states that have the least negative reward
amongst its immediate neighbors.  Given the adjacency matrix $A$, it is then at state $s$, an intuitive model for the agent's actions can be
\begin{align}
w_r(s,s')=\frac{\exp^{\beta R(s')}}{\sum_{s\sim s''} {\exp}^{\beta R(s'')}},
\end{align}
where $\beta>0$ is a positive constant that models affinity. In this paper, we construct the Reward based Proto-Value Functions (RPVFs) by looking at the $n\times n$ diffusion matrix $W_R=(w_r(s,s'),s\in S,s'\in S)$ which combines the task-dependent rewards and the task-independent connectivity information.\par

\section{Experiments}
We demonstrate the following via the experiments in this section.\\
$1)$ \textbf{Similarity matrices other than the diffusion matrix can be used to generate features:} To this end, show that the Gaussian kernel matrix generated using the optimal value function as data points also yields good features. Further, we show that the proto-value functions of the \emph{three-room} problem \cite{icmlproto} can be recoverd even when the walls are absent if one assigns appropriate negative rewards for those cells corresponding to the `wall' states. In short, we show that using $W$ or $W_R$ (with negative rewards) is equivalent in this case.\\
$2)$ \textbf{Reward shaping does not work with all the features:} We show that irrespective of whether additional reward shaping is used or not, the profile of the learnt value function is limited to the choice of the basis. In particular, when the state space is symmetrical and the rewards are asymmetrical, the RPVF capture the asymmetry better than the PVFs.\\
\textbf{The Gaussian Kernel}\par Given a set $\{x_1,\ldots,x_n\}\subset \R^d$ of $n$ data points in $d$-dimensions, the $n\times n$ \emph{Gaussian} kernel matrix $\K=(K(i,j))$ is given by 
\begin{align}\label{gauss}
K(x_i,x_j)=exp^{-\frac{||x_i-x_j||}{2\sigma^2}},
\end{align}
where $\sigma>0$ is a positive scaling constant.
Note that $\K$ is a similarity matrix which assigns the \emph{nearby} states a higher value, a fact evident from \eqref{gauss}. The spectral clustering technique involves computing the top $k$ eigen-vectors of $\K$ to obtain a $k$-dimensional embedding $\{y_i,i=1\ldots,n\}\subset \R^k$, where $y_i=(y_i(1),\ldots,y_i(k))\in \R^k$ with $y_i(j)$ being the $i^{th}$ component of the $j^{th}$ eigen-vector. In an MDP, we are interested in the set of data points $\{J^*(1),\ldots,J^*(n)\}\subset\R$ and the kernel matrix with entries
\begin{align}\label{gaussval}
K(x_i,x_j)=exp^{-\frac{||J^*(i)-J^*(j)||}{2\sigma^2}}.
\end{align}
We observe that the second eigen-vector of the kernel matrix in \eqref{gaussval} is a close approximation to the optimal value function.
\begin{figure}
	\begin{center}
		\includegraphics{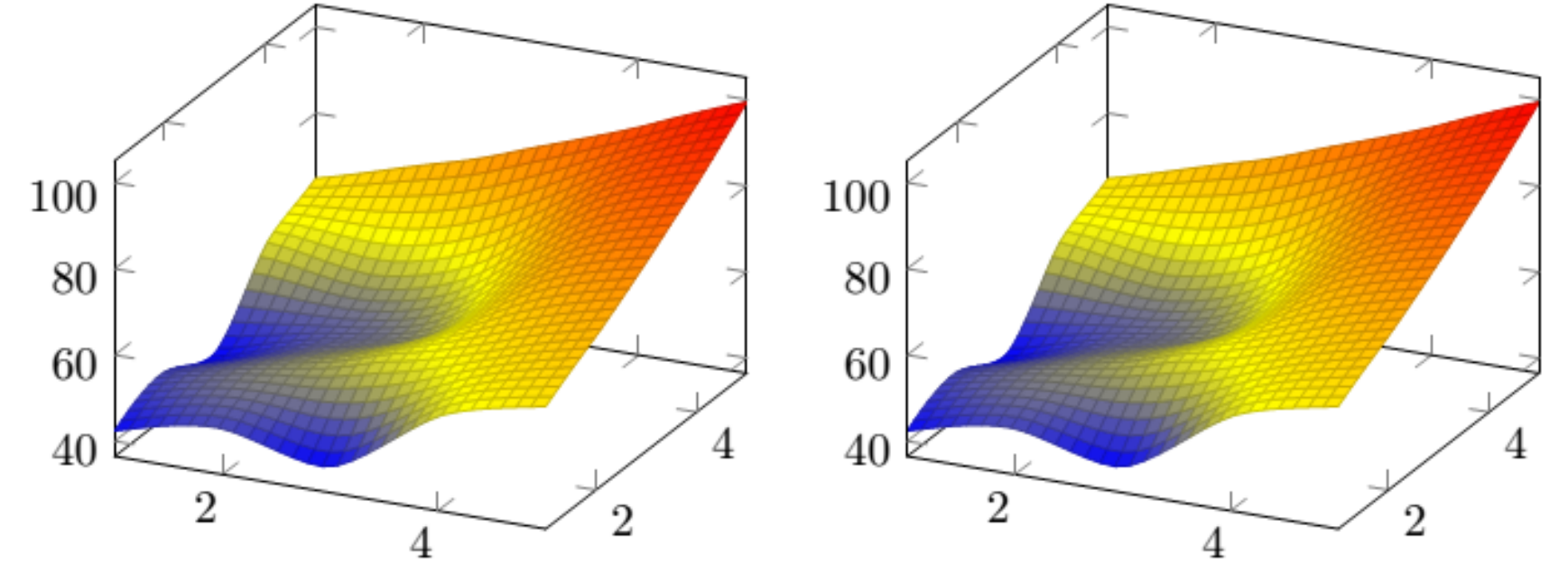}
		\caption{On the left is the optimal value function and on the right is the first eigen function of the matrix $K$ generated using $J^*$ and $\sigma=0.1$.}
		\label{threeroomvalproto}
	\end{center}
	
\end{figure}
The eigen-vectors of the graph Laplacian of the $3$-room MDP and the graph Laplacian of the reward based diffusion matrix $W_R$ of a simple $21\times 60$ grid which has negative rewards in the place of the wall are shown in \Cref{threeroomvalproto}.\\
\begin{figure}
	\begin{center}
		\includegraphics{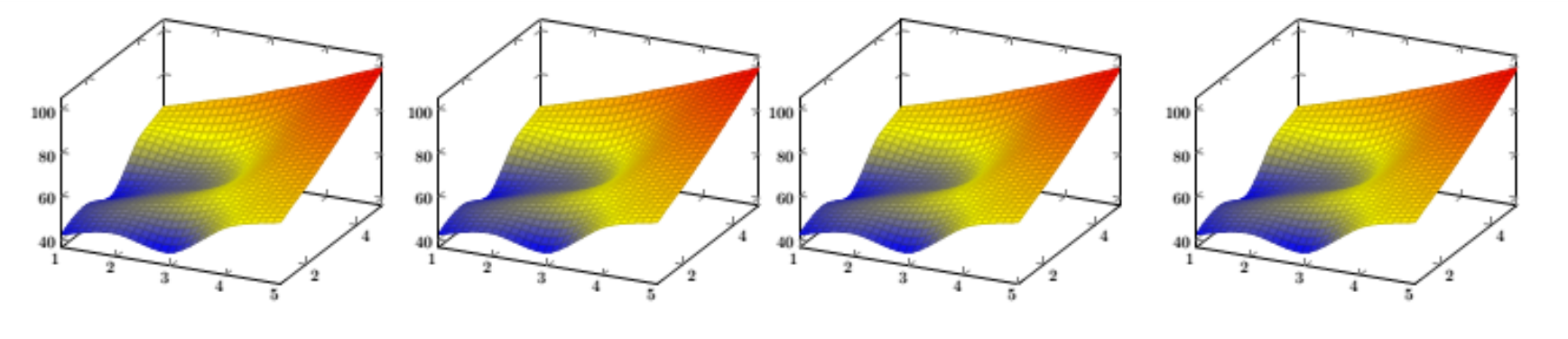}
		\caption{First two on the left  are the eigen functions of the $W$ matrix and two on the right are the eigen functions of the $W_R$ ($\beta=0.1$) matrix.}
		\label{threeroomrewprot}
	\end{center}
	
\end{figure}
\begin{table}[h!]
    \begin{minipage}{.5\linewidth}
      
      \centering
        \begin{tabular}{|c|c|c|c|c|}\hline
            \mb& \mb& \mb& \mb& $\mathbf{G}$\\\hline
            \mb& \mb& $\uparrow$& \mb& \mb\\\hline
            \mb& $\leftarrow$& \mb& $\ra$& \mb\\\hline
            \mb& \mb& $\downarrow$& \mb& \mb\\\hline
            \mb& \mb& \mb& \mb& \mb\\\hline
        \end{tabular}
\label{gwprb}
    \end{minipage}%
    \begin{minipage}{.5\linewidth}
      \centering
        \begin{tabular}{|c|c|c|c|c|}\hline
            5& 10& 15& 20& 25\\\hline
            4& 9& 14& 19& 24\\\hline
            3& 8& 13& 18& 23\\\hline
            2& 7& 12& 17& 22\\\hline
            1& 6& 11& 16& 21\\\hline
        \end{tabular}
        
\label{rshaping}
    \end{minipage} 
    \caption{On the left is the grid world task with a reward of $10$ in the goal-state $\mathbf{G}$. On the right is the potential reward shaping function according to \Cref{th}.}
\label{rshape}
\end{table}

\par
\textbf{Does Reward shaping work with any features?}
We now look at an instance of goal-based MDP (see \Cref{rshape}). Here, the goal-state is in the right hand corner. The agent receives a reward of $10$ on reaching the goal state and actions in the intermediate states do not receive any reward. The allowable actions are to move \emph{up, down, right} or \emph{left}. The state space for an $N\times N$ grid (such as the one in \Cref{rshape}) is given by $S=\{s=(x,y),x=1,\ldots,N, y=1,\ldots,N\}$, where $(1,1)$ denotes the \emph{bottom-left} cell and $(N,N)$ the \emph{top-right} cell. It is evident that the learning process can be sped if the agent is rewarded for those actions that take it either \emph{up} or \emph{right}.
\begin{align}
\psi(x,y)>\psi(x',y,'), \forall x>x', y>y'.
\end{align}
right table in ). Even in this case the profile of the learnt value function did not change, and the resulting policy performed only moderately.
\begin{figure}
	\begin{center}
		\includegraphics{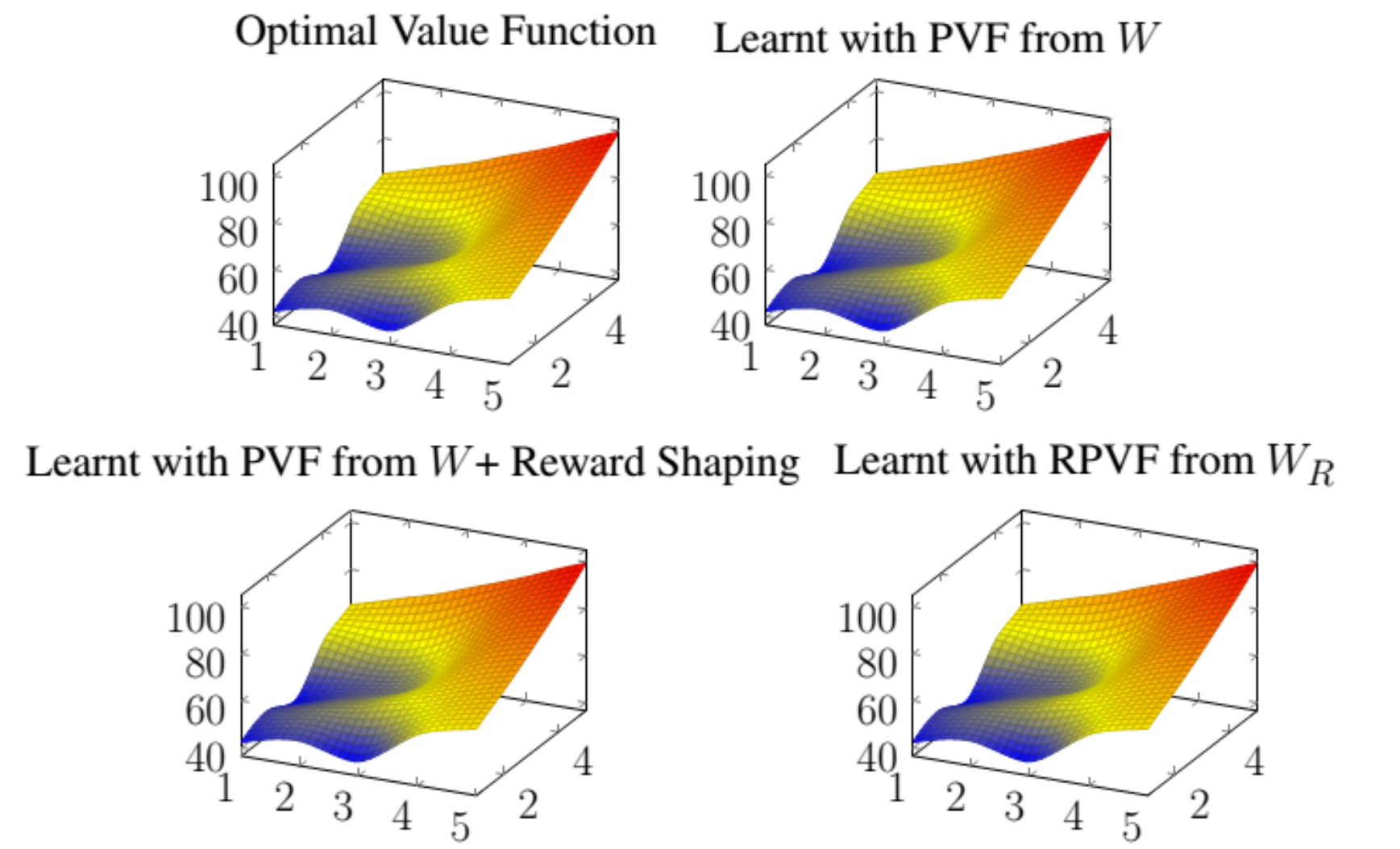}
		\caption{Shows the optimal value function (top most) and the learned value functions (bottom three) for the grid world domain in. Notice that learning using the RPVF (bottom most) is better than the learning using PVF (middle two) with/without reward shaping.}
		\label{gplt}
	\end{center}
	
\end{figure}
We ran the RPI algorithm with $\Theta=W_R$ (with $\beta=1$) and the results are shown (second from bottom) in . In this case, the profile of the learnt value function resembles the optimal value function. Further, we also observed that the policy $\pi_{W_R}$ returned by RPI in this case performed better than $\pi_{W}$ (with/without reward shaping), i.e., $\sum_{s\in S} J_{\pi_{W_R}}(s)=1660$.\par
The reason why the RPVFs perform better than the PVFs can be explained by looking at the corresponding eigen functions.


\begin{figure}
	\begin{center}
		\includegraphics{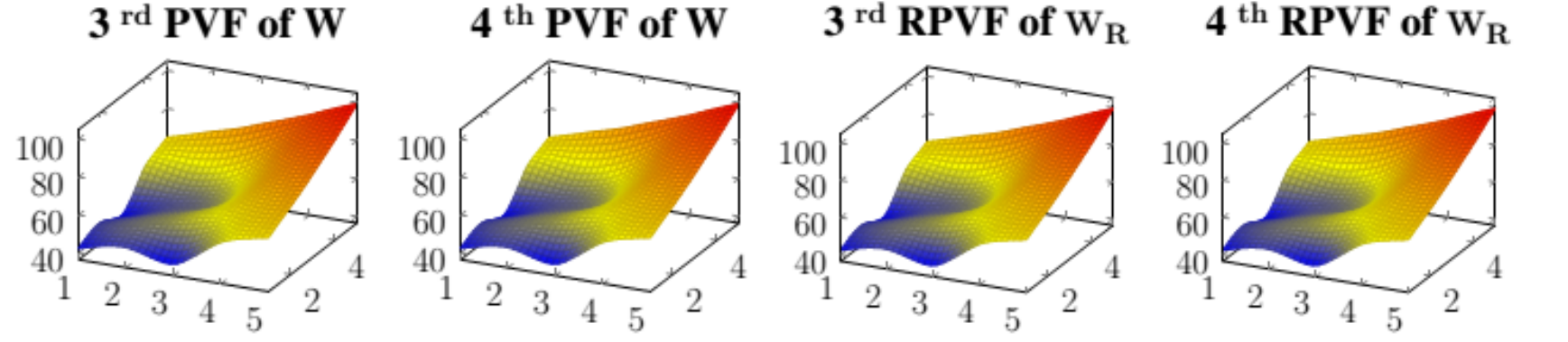}
		 \caption{Comparing the profiles of eigen functions of $W$ and $W_R$. The first two eigen functions of $W$ and $W_R$ were identical and hence not presented.}
		 \label{compare}
	\end{center}
	
\end{figure}
We made use of the PVF based representation for the grid world problem in \Cref{rshape}. The optimal value function is shown in the top most plot of \Cref{gplt}
We chose $k=4$, i.e., $4$ eigen functions corresponding to $4$ largest eigen values of the diffusion matrix $\Ps$ constructed from the adjacency matrix. We ran the RPI algorithm with $\Theta=W$ and the result is shown (second from top) in \Cref{gplt}. Notice that the value function learnt by the RPI algorithm does not quite resemble the profile of the optimal value function and consequently resulted only in a moderately good policy. We evaluated the policy $\pi_{W} $returned by RPI in this case (i.e., $\Theta=W$) and it turned out that $\sum_{s\in S}J_{\pi_{W}}(s)=1132$ as opposed to $\sum_{s\in S}J^*(s)=1887$. Further, we also ran the RPI, by retaining $\Theta=W$, however provided additional reward shaping feedback using the potential function $\psi$ (see  
\par The eigen functions corresponding to the first two largest eigen values were the same in both the cases. However, the third and fourth eigen functions differed (see \Cref{compare}). We can see from (bottom most plot) \Cref{gplt}, that this difference in eigen function shows up in the difference in the profiles of the corresponding learnt value functions.     
\begin{table}[!htb]

    \begin{minipage}{.5\linewidth}
      
      \centering
        \begin{tabular}{|c|c|c|c|c|}\hline
            0& 0& 0& 0& 10\\\hline
            0& 0& 0& 0& 0\\\hline
            {\color{red}\ding{53}}& 0& {\color{red}\ding{53}}& {\color{red}\ding{53}}& 0\\\hline
            0& 0& 0& {\color{red}\ding{53}}& 0\\\hline
            0& 0& {\color{red}\ding{53}}& 0& 0\\\hline
        \end{tabular}
    \end{minipage}%
%
%
    \caption{On the left is the grid world task with mines (marked as {\color{red}\ding{53}}). On visiting a mine state the agent receives a random reward between $-1$ to $-5$.}
\end{table}

We also compared the performance of PVFs and RPVFs in a variant of the grid world problem, where, in addition to the goal-state, there are certain \emph{mine} states with negative rewards. We chose these mine states at random and then compared the performances across $10$ such different random grid world problems and for each problem we averaged the result across $10$ initial policies for the RPI. We observed that in $9$ out of the $10$ systems, RPI with RPVF features (generated for $\beta=0.1$) significantly outperforms the policy learnt using the PVF.
\begin{figure}[h!]
\centering

\resizebox{\columnwidth}{!}
{
\tiny
\begin{tikzpicture}
\begin{axis}[
    ybar,
    bar width=0.2,
    legend style={at={(0.5,-0.15)},
      anchor=north,legend columns=-1},
    ylabel={$\sum_{s\in } J_{\pi_{\Theta}}(s)$},
    symbolic x coords={S1,S2,S3,S4,S5,S6,S7,S8,S9,S10},
    nodes near coords align={vertical},
    ]
\addplot coordinates {(S1,1096) (S2,830) (S3,1051) (S4,1008) (S5,885) (S6,1015) (S7,1011) (S8,1022) (S9,1014) (S10,1022) };
\addplot coordinates {(S1,1628) (S2,1150) (S3,1530) (S4,1361) (S5,1454) (S6,1486) (S7,1496) (S8,1502) (S9,822) (S10,1504) };
\legend{PFV,RPFV,not understood}
\end{axis}
\end{tikzpicture}
}
\label{barplot}
\caption{Compares the performance of RPVFs vs PVFs on $10$ different \emph{mine}-grid tasks. Here the performances have been averaged over $10$ different initail policies. Here $\pi_{\Theta}$ is the policy returned by RPI, with $\Theta=W$ and $\Theta=W_R$ for PVFs and RPVFs respectively.}
\end{figure}
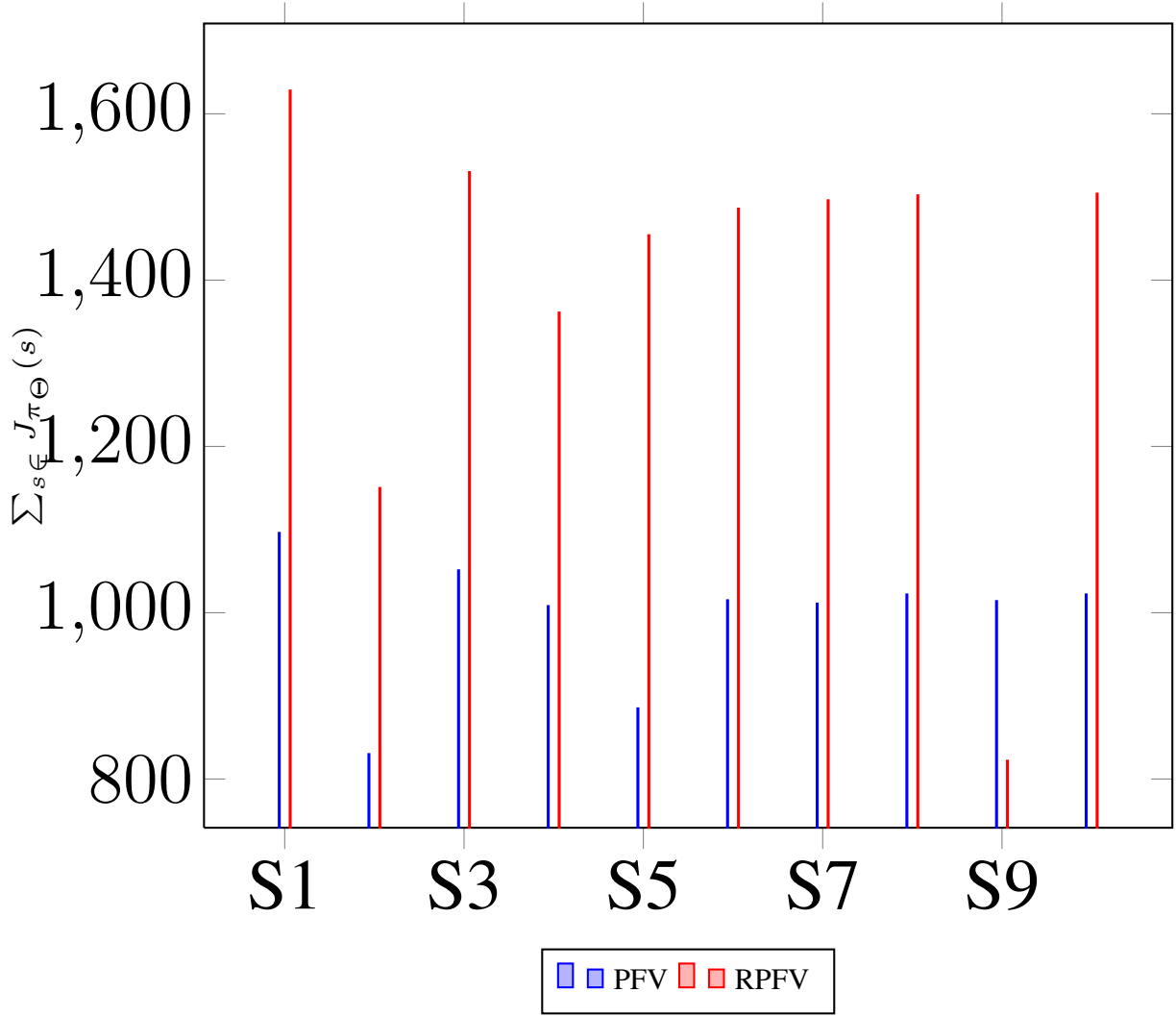

\vspace{-10pt}
\section{Conclusion}
We combined the task-independent proto-value function (PVF) construction and the task-specific reward shaping to obtain Reward based Proto-Value Functions (RPVFs). The RPVF construction made use of the immediate rewards which were avaialble during the sampling phase but were not used in the PVF construction.
We also observed that the RPVFs perform better than the PVFs in goal-based RL tasks. The salient feature of the RPVFs was that captured the asymmetry in the value function induced by the reward structure better than the PVFs. As an interesting future direction, we can look at extending RPVF to continous domains.

\bibliographystyle{plain}
\bibliography{ref.bib}
\newpage
\onecolumn
\end{document}